\documentclass[twoside]{article}
\usepackage[preprint]{aistats2027}
\usepackage[T1]{fontenc}
\usepackage[utf8]{inputenc}
\usepackage{amsmath,amssymb,graphicx,booktabs,tabularx,microtype,placeins}
\usepackage[round]{natbib}
\usepackage[hidelinks]{hyperref}
\hypersetup{
  pdftitle={Empirical Evidence for Simply Connected Decision Regions in Image Classifiers},
  pdfauthor={Arjhun Swaminathan and Mete Akgün}
}

\begin{document}
\runningtitle{Simply Connected Decision Regions in Image Classifiers}
\runningauthor{Arjhun Swaminathan, Mete Akg\"un}
\twocolumn[
\aistatstitle{Empirical Evidence for Simply Connected Decision Regions in Image Classifiers}
\aistatsauthor{Arjhun Swaminathan \And Mete Akg\"un}
\aistatsaddress{Medical Data Privacy and Privacy-preserving Machine Learning (MDPPML) \\
Institute for Bioinformatics and Medical Informatics (IBMI) \\
University of T\"ubingen, Germany \\
\texttt{\{arjhun.swaminathan, mete.akguen\}@uni-tuebingen.de}}
]

\begin{abstract}
The topology of a classifier's decision regions determines how inputs with the same predicted label can be connected and deformed without changing that prediction. Prior empirical work constructed paths between same-label images within a single region, but did not examine whether loops bound surfaces within that region. We investigate this question using adaptive quadrilateral meshes with targeted repair of off-label interior vertices, while holding the same-label boundary loop fixed. A finite-resolution acceptance criterion distinguishes completed constructions from those left unresolved at the refinement ceiling. Across the pretrained classifiers studied, every tested loop admits an accepted filling. Construction effort varies by orders of magnitude within classes and is greater for mean-score-adjusted randomly initialised classifiers than for trained classifiers. An analytic control with a known hole leaves winding loops unresolved at the tested hole radii at or above the resolution threshold. These results provide empirical evidence consistent with simply connected decision regions at the tested resolution.
\end{abstract}

\section{Introduction}
Deep neural networks have achieved remarkable performance in image classification, with modern architectures ranging from convolutional networks to vision transformers \citep{he2016deep,tan2019efficientnet,huang2017densely,liu2022convnet,dosovitskiy2020image,liu2021swin,krizhevsky2012imagenet,simonyan2014very,szegedy2015going}. Despite their success, the geometry of the decision regions learned by these models remains incompletely understood \citep{ramamurthy2019topological,fawzi2018empirical}. A classifier partitions the input space into regions corresponding to predicted labels. Their topology describes whether images of the same predicted class can be joined within that class and whether loops can be contracted without leaving it. Their boundary geometry describes a different, local question: how close label-changing inputs lie to natural images. Understanding both properties is relevant to interpretability and adversarial analysis \citep{fawzi2018empirical,goodfellow2014explaining,szegedy2013intriguing,moosavi2016deepfool,moosavi2017universal}. Establishing which global properties these regions possess, and how they differ across training conditions, is therefore a fundamental question about classifier structure.

A central observation motivating this work is that local adversarial vulnerability does not necessarily imply global topological fragmentation. Although small perturbations can move an image across a decision boundary, the corresponding decision region may still be large and path connected. Prior empirical work showed that, for several deep image classifiers, pairs of images assigned to the same class could often be connected by continuous, nearly straight paths within the same decision region \citep{fawzi2018empirical}. Decision-boundary geometry has also informed adversarial detection and predicted-label-based attacks \citep{fawzi2018empirical,brendel2017decision,chen2020hopskipjumpattack}; these applications motivate geometric analysis without following from connectivity alone.

This path-connectivity perspective opens a direction for studying neural networks in input space. A successful path between two images shows that they lie in the same path-connected component, but does not rule out non-contractible loops in that component. The next question is whether a loop can be extended to a continuous map of a disk within the decision region \citep{hatcher2005algebraic}. A two-dimensional filling is dictated by this topological question, rather than used to approximate the ambient high-dimensional space. The distinction also has a constructive interpretation: two paths with common endpoints form a loop when one is traversed in reverse, and a continuous label-preserving filling would allow one path to be deformed into the other with endpoints fixed. At finite resolution, inspecting candidate surfaces provides an empirical way to compare interpolation routes and visualise two-parameter families of images.

Surface construction has a long history in geometric modelling \citep{barnhill1977representation}. In classical computer-aided design, Coons patches \citep{coons1967surfaces} provide a principled way to construct a surface from four boundary curves by blending the boundaries and correcting for the bilinear interpolation of the vertices. Given four same-label paths forming a closed loop, the Coons construction supplies a canonical candidate surface. Its interior, however, need not remain in the same decision region.

In this paper, we study loop fillability and construction effort using adaptive quadrilateral meshes and targeted vertex repair. Four same-label boundary paths are held fixed during interior filling, and the returned patches are checked under a specified grey-RMS resolution and refinement budget. We evaluate $1000$ configurations for each of six ImageNet classifiers spanning convolutional and transformer architectures, namely ResNet-50, DenseNet-121, EfficientNet-B0, ConvNeXt-Tiny, ViT-B/16 and Swin-T; all $6000$ satisfy the stated finite-resolution acceptance criterion. A CIFAR-100 comparison then examines the effort required for trained classifiers and mean-score-adjusted randomly initialised classifiers, using mesh size, repair count and unresolved outcomes. Repeated-loop and anchor-separation studies assess sensitivity to the sampled configurations. The default threshold is half of one $8$-bit grey level per pixel, and in a control region carrying a known hole, a winding loop is left unresolved by this same procedure at the tested hole radii at or above the threshold (Section~\ref{sec:results}). Every tested pretrained-classifier configuration is accepted under this criterion. We therefore report empirical evidence for loop fillability at the tested finite resolution, consistent with these decision regions being not only path connected, as earlier work showed \citep{fawzi2018empirical}, but also simply connected at that resolution, subject to the sampling qualifications in Section~\ref{sec:setup} and Appendix~\ref{app:sampling}. To our knowledge this is the first empirical study of loop fillability in the decision regions of deep image classifiers.

Section~\ref{sec:related} reviews related work; Sections~\ref{sec:method} and~\ref{sec:setup} describe the method and experimental setup. Section~\ref{sec:results} presents results, including the known-hole control, followed by interpretation and limitations in Section~\ref{sec:discussion} and conclusions in Section~\ref{sec:conclusion}.

\section{Related work}
\label{sec:related}
The topology of neural-network decision regions has been studied theoretically and empirically. Under their respective architectural and activation assumptions, prior results relate network width to the ability to represent disconnected regions and to the boundedness of their connected components \citep{nguyen2018neural,beise2021decision}, and depth has been related to the attainable complexity of the regions themselves \citep{bianchini2014complexity}. These results establish that architecture can constrain region topology; they do not establish simple connectedness for the classifiers considered here. Persistent homology provides a complementary approach through sampled geometric complexes \citep{edelsbrunner2010computational,ghrist2008barcodes,carlsson2009topology}. In particular, labelled complexes have been used to infer the homology of decision boundaries and to support model selection \citep{ramamurthy2019topological}, while other work studies internal representations \citep{watanabe2022topological}. Topological structure has also been applied outside classifier analysis, for example to bound re-identification risk across sequential data releases \citep{swaminathan2025dynamic}. Our construction instead produces explicit patches in image space and measures the effort required to obtain them.

Decision-boundary geometry also informs adversarial analysis. DeepFool uses local linearisation to estimate label-changing perturbations \citep{moosavi2016deepfool}; decision-based attacks search using predicted labels alone, with later work reducing the query budget needed in the targeted setting \citep{brendel2017decision,chen2020hopskipjumpattack,swaminathan2025accelerating}, and a line of work exploits the local shape of the boundary itself, through its normal directions, its low mean curvature or a surrogate-free surface search \citep{rahmati2020geoda,maho2021surfree,reza2023cgba}. These methods estimate where the boundary lies near a given image; we instead ask whether a loop inside one region bounds a surface, which is a global question about the region and not about the boundary near a point. We use a targeted DeepFool-style repair and therefore require access to classifier scores and gradients, even though the decision regions and acceptance tests are defined by hard labels.

The closest empirical predecessor constructed paths between same-label images and studied local decision-boundary geometry \citep{fawzi2018empirical}. We extend the construction problem from paths to two-dimensional patches. The comparison concerns the topology of a loop, rather than a low-dimensional representation of the entire input space. The Coons construction supplies a boundary-matched geometric reference \citep{coons1967surfaces}.

\section{Method}
\label{sec:method}
\subsection{Problem setup}
\label{subsec:problem_setup}
We evaluate the classifier on real-valued normalised image arrays $\mathcal{X}=\mathbb{R}^{3HW}$. Interpolated and repaired arrays need not remain on a natural-image manifold or within the raw $8$-bit pixel range. Let $f:\mathcal{X}\to\{1,\ldots,K\}$ be the predicted-label classifier and let $\mathcal{R}_y=\{x\in\mathcal{X}:f(x)=y\}$ be its decision region for label $y$. Four images $x_{00},x_{10},x_{01},x_{11}\in\mathcal{R}_y$ define the anchor order $x_{00}\to x_{10}\to x_{11}\to x_{01}\to x_{00}$.

The ideal topological target is a continuous map $S:[0,1]^2\to\mathcal{R}_y$ with a specified boundary loop. Its four same-label boundary paths, obtained by the path construction of \citet{fawzi2018empirical}, are held fixed during surface filling. Refinement samples these curves without altering them; off-label interior vertices may be moved by repair. Labels are verified at stored vertices and grid samples of each patch.

Let $\gamma_0(u)$, $\gamma_2(u)$, $\gamma_3(v)$ and $\gamma_1(v)$ denote the fixed bottom, top, left and right boundary paths, respectively, with endpoints matched to the anchors. The boundary conditions are
\begin{equation}
\begin{aligned} S(u,0)&=\gamma_0(u), & S(u,1)&=\gamma_2(u),\\ S(0,v)&=\gamma_3(v), & S(1,v)&=\gamma_1(v). \end{aligned}
\end{equation}

\subsection{Resolution and reference surface}
Let $\sigma_c$ be the channel standard deviations used in the model's input normalisation. Following the implementation, we express distance on a grey-level RMS scale using
\begin{equation}
\begin{aligned}
\ell_{\mathrm{grey}}&=\sqrt{HW\sum_{c=1}^{3}\left(\frac{1}{255\sigma_c}\right)^2},\\
 d_{\mathrm{grey}}(x,x')&=\frac{\|x-x'\|_2}{\ell_{\mathrm{grey}}}.
\end{aligned}
\end{equation}
For raw grey-level values $p,p'$, this is the normalisation-weighted RMS quantity
\begin{equation}
d_{\mathrm{grey}}^2(x,x')=
\frac{\sum_{c,h,w}(p_{chw}-p'_{chw})^2/\sigma_c^2}
{HW\sum_c1/\sigma_c^2}.
\end{equation}
It equals one for a one-grey-level change at every pixel and channel. For general perturbations it need not equal the unweighted pixel RMS. We use the dataset-specific normalisation in each experiment and define the stopping threshold in this metric.

Given four boundary curves, the Coons reference is
\begin{equation}
\begin{aligned} C(u,v)={}&(1-v)\gamma_0(u)+v\gamma_2(u)\\ &+(1-u)\gamma_3(v)+u\gamma_1(v)-B(u,v). \end{aligned}
\end{equation}
where the bilinear corner interpolation is
\begin{equation}
\begin{aligned} B(u,v)={}&(1-u)(1-v)x_{00}+u(1-v)x_{10}\\ &+(1-u)vx_{01}+uvx_{11}. \end{aligned}
\end{equation}
\begin{figure}[t]
\centering
\includegraphics[width=\linewidth]{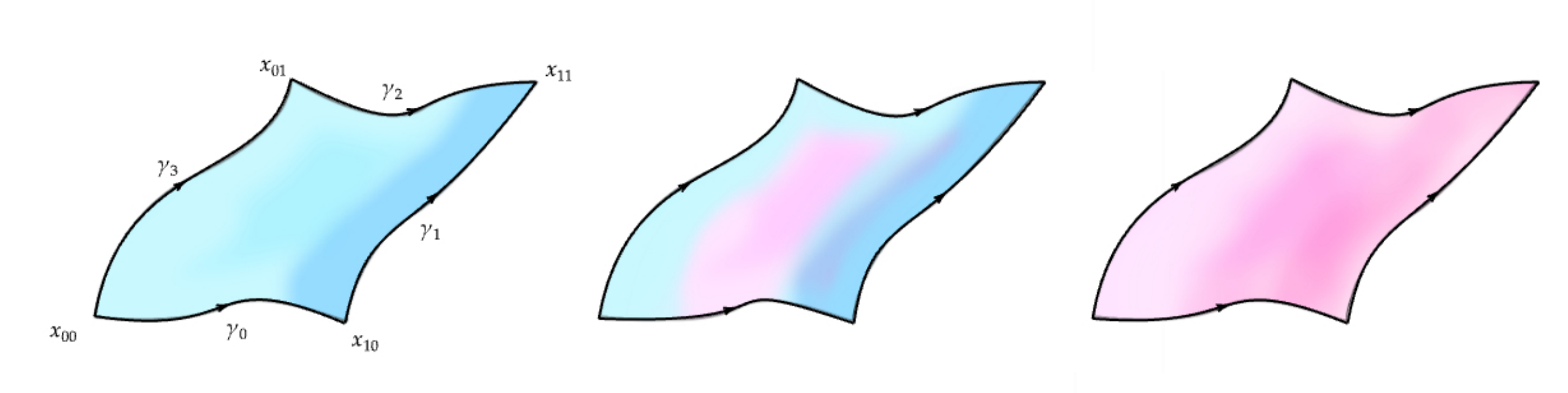}
\caption{Two-dimensional illustration of the surface-filling construction. Left: Coons reference induced by the boundary curves. Middle: overlay of the reference and a candidate filling. Right: candidate filling.}
\label{fig:coons_overlay}
\end{figure}

The initial root patch uses the bilinear corner interpolation $B$; Figure~\ref{fig:coons_overlay} illustrates the reference and a candidate filling in two dimensions. For straight boundary curves it coincides with the Coons patch. The boundary curves remain fixed throughout the filling procedure, while interior vertices introduced by refinement may subsequently be repaired. The filling procedure does not repeatedly project an entire Coons patch into the target region.

\subsection{Subdivision, repair and acceptance}
The mesh uses dyadic parameter coordinates and a shared registry of image-space vertices. A quad with corners $z_{00},z_{10},z_{01},z_{11}$ represents the local bilinear patch
\begin{equation}
\begin{aligned} Q(u,v)={}&(1-u)(1-v)z_{00}+u(1-v)z_{10}\\ &+(1-u)vz_{01}+uvz_{11}. \end{aligned}
\end{equation}
Its grey-RMS diameter is the largest $d_{\mathrm{grey}}$ distance between its four vertices. A patch is accepted by size when all four vertices have the target prediction and this diameter is at most $\tau$. Such patches are not subjected to an interior label grid.

Larger patches are checked on $G_m=\{(a/m,b/m):0\leq a,b\leq m\}$. A patch is grid-accepted when every evaluated point has label $y$. The implementation chooses a common grid size for the active larger patches at each level. With base grid size $b$, level $d$, minimum grid size $m_{\min}$ and a maximum diameter $\widehat\Delta_d$ estimated from at most $256$ active patches, it sets
\begin{equation}
\begin{aligned} r_d&=\max\{m_{\min},\lceil b/2^d\rceil,\lceil\widehat\Delta_d/\tau\rceil\},\\ m_d&=\min\{m_{\max},2^{\lceil\log_2 r_d\rceil}\}. \end{aligned}
\end{equation}
The grid cap and the sampled diameter estimate make this a sampling rule, not a guarantee that every neighbouring pair is separated by at most $\tau$.

\begin{figure*}[t]
\centering
\includegraphics[width=\linewidth]{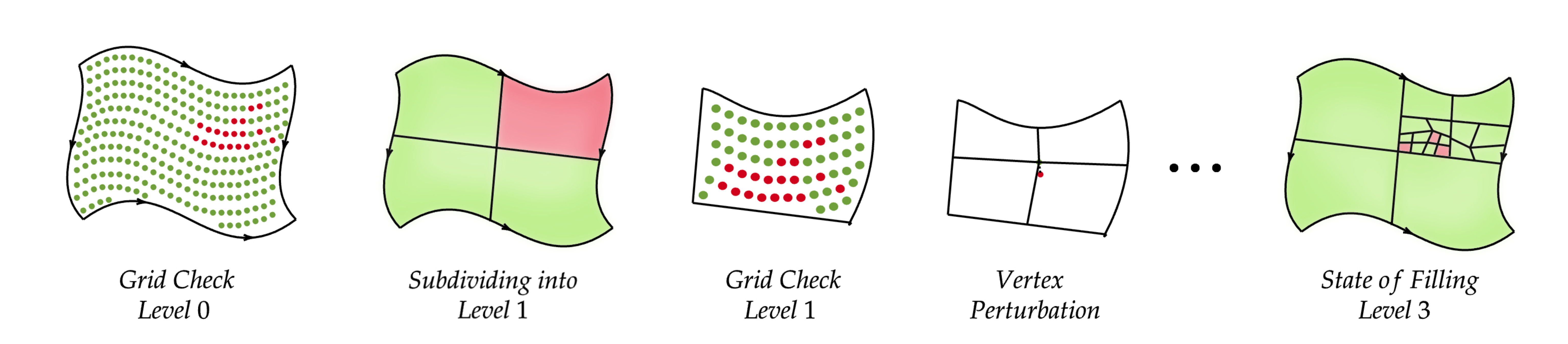}
\caption{Illustration of the adaptive surface-filling procedure. Green and red points indicate samples with and without the target label. A patch that fails the grid check is subdivided into four quads. Newly introduced off-label vertices are repaired by an approximate local update.}
\label{fig:filling_procedure}
\end{figure*}

A failing patch is subdivided into four child quads. Interior edge midpoints are initialised by averaging their endpoints, and the centre by averaging the four corners. New vertices lying on the boundary are obtained from the corresponding fixed boundary curve and are not moved by interior repair. Newly introduced interior vertices whose prediction differs from $y$ are updated by targeted DeepFool-style repair, with an amplification fallback if needed, followed by target-side bisection. The replacement is checked for label $y$; it is not an exact nearest-point projection onto $\mathcal{R}_y$. Repair failure is reported separately from a completed mesh. Figure~\ref{fig:filling_procedure} shows one level of this cycle.

We call a construction accepted only when a mesh is returned, the run completes successfully, and no patch is accepted solely because the maximum parameter-refinement depth was reached; a frontier left at that ceiling makes the construction unresolved. The distinction matters: a patch accepted by size meets the stated geometric criterion, whereas one accepted at the ceiling meets no criterion at all.

Construction cost is dominated by classifier evaluations rather than by mesh bookkeeping. A grid check at level $d$ costs $(m_d+1)^2$ forward evaluations for each active patch, a failed check replaces that patch by four children, and each repaired vertex uses at most $50$ gradient steps, up to $12$ amplification evaluations if needed, and $10$ bisection evaluations. The depth budget bounds the number of levels, but the levels actually used depend on the construction. We report final quad count, repair count and wall-clock time (Appendix~\ref{app:provenance}). Memory is dominated by the shared vertex registry, which stores one image array per distinct mesh vertex.

Shared vertices make equal-level edges agree, while repaired hanging midpoints can mismatch where refinement levels differ. Appendix~\ref{app:visual} measures these mismatches directly and finds them at most $1.1\times10^{-6}\tau$, alongside a dense-label audit. Acceptance throughout means the stated empirical criterion, verified at sampled points rather than along a continuum.

\section{Experimental setup}
\label{sec:setup}
\paragraph{Datasets and classifiers.}
We test breadth on ImageNet validation images across six pretrained architectures: ResNet-50, DenseNet-121, EfficientNet-B0, ConvNeXt-Tiny, ViT-B/16 and Swin-T \citep{deng2009imagenet,russakovsky2015imagenet}. For a training comparison we use CIFAR-100 with a compact VGG-style convolutional network with batch normalisation \citep{krizhevsky2009learning}, trained from two seeds that reach test accuracies $0.6779$ and $0.6829$ at epoch $30$.

\paragraph{Anchor selection.}
Each ImageNet architecture contributes $1000$ configurations, one per predicted label. For a given label we take the first four images the model assigns to it, deterministically rather than by a uniform random draw, and an image qualifies only if its predicted label is unchanged after it is written to disk and read back, which is how the experiment stores its anchors. Sixty-one configurations were completed after the initial run, by substituting same-class anchors or by rerunning a missing output; Appendix~\ref{app:provenance} gives the breakdown. On CIFAR-100 we instead draw ten distinct four-anchor cycles from six stable images in each eligible class, giving $1000$ loops for each trained seed.

\paragraph{The randomly initialised comparison.}
A raw randomly initialised network predicts one class for almost every input, leaving no diversity of regions to test. We therefore subtract each logit's fixed mean over the test pool before taking the predicted label, and use this adjusted classifier for both anchor selection and filling. It leaves $77$ and $75$ eligible classes for the two seeds, so these arms contain $770$ and $750$ loops; the shortfall reflects classes lacking six stable images, rather than truncation of a larger set. The adjustment also changes the classifier, so this comparison does not isolate the causal effect of training alone. Appendix~\ref{app:sampling} gives the protocols and the raw-initialisation diagnostic.

\paragraph{Construction settings.}
All principal comparisons use $\tau=0.5$, refinement depth $14$, base grid size $100$, $m_{\min}=2$ and $m_{\max}=512$, with repair limited to $50$ DeepFool iterations, overshoot $0.02$ and $10$ bisection steps; batch sizes vary with the memory each model requires. We measure construction effort by final mesh size and vertex-repair count under these settings. Root acceptance means the initial bilinear patch passes without subdivision, and the initial-patch on-label fraction is a separate finite-grid diagnostic; neither quantity measures global convexity.

\section{Results}
\label{sec:results}
\subsection{Results across architectures}
Every one of the $6000$ ImageNet constructions satisfies the acceptance criterion under default repair, as Table~\ref{tab:imagenet} reports. Most configurations require subdivision beyond the initial bilinear patch, and root acceptance ranges from $6.5\%$ to $41.5\%$ across architectures. This describes the sampled configurations and the interpolation rule, not the fraction of a class region that is convex.

\begin{table*}[t]
\centering
\small
\caption{ImageNet construction outcomes under default repair, over $1000$ configurations per model. Root acceptance is the proportion accepted without subdivision; mean quad count includes the same configurations.}
\label{tab:imagenet}
\begin{tabular}{lrrrr}
\toprule
Model & Tested & Accepted & Root (\%) & Mean quads \\
\midrule
ResNet-50 & 1000 & 1000 & 15.9 & 15644.1 \\
DenseNet-121 & 1000 & 1000 & 6.5 & 24248.0 \\
EfficientNet-B0 & 1000 & 1000 & 8.4 & 21188.4 \\
ConvNeXt-Tiny & 1000 & 1000 & 38.6 & 5197.5 \\
ViT-B/16 & 1000 & 1000 & 27.9 & 6799.2 \\
Swin-T & 1000 & 1000 & 41.5 & 4462.1 \\
\bottomrule
\end{tabular}
\end{table*}

Root acceptance and mean quad count rank the six architectures in exactly opposite orders, with Spearman correlation $-1$ between them (Table~\ref{tab:imagenet}). The three more recent architectures, ConvNeXt-Tiny, ViT-B/16 and Swin-T, sit at the inexpensive end and the three earlier convolutional networks at the expensive end, separated by roughly a factor of six in root acceptance and five in mesh size. We report this ordering without an architectural explanation. Each model contributes its own anchor configurations, input normalisation and class regions under the deterministic selection rule, so this is not a controlled architecture experiment, and Appendix~\ref{app:curvature} shows that the accompanying geometric diagnostics do not support reading such an ordering as an intrinsic property of the regions.

No construction in this collection approached the refinement ceiling; the deepest reached level $10$ of a possible $14$. The spread of effort is wide even so, as the easy, median and hard examples of Figure~\ref{fig:animals} show, drawn from the separate anchor-separation study of Appendix~\ref{app:distance}. Throughout we use final quad count $Q$ as the primary measure of construction effort, with refinement depth and vertex-repair attempts as complementary diagnostics.

\begin{figure*}[t]
\centering
\includegraphics[width=\linewidth]{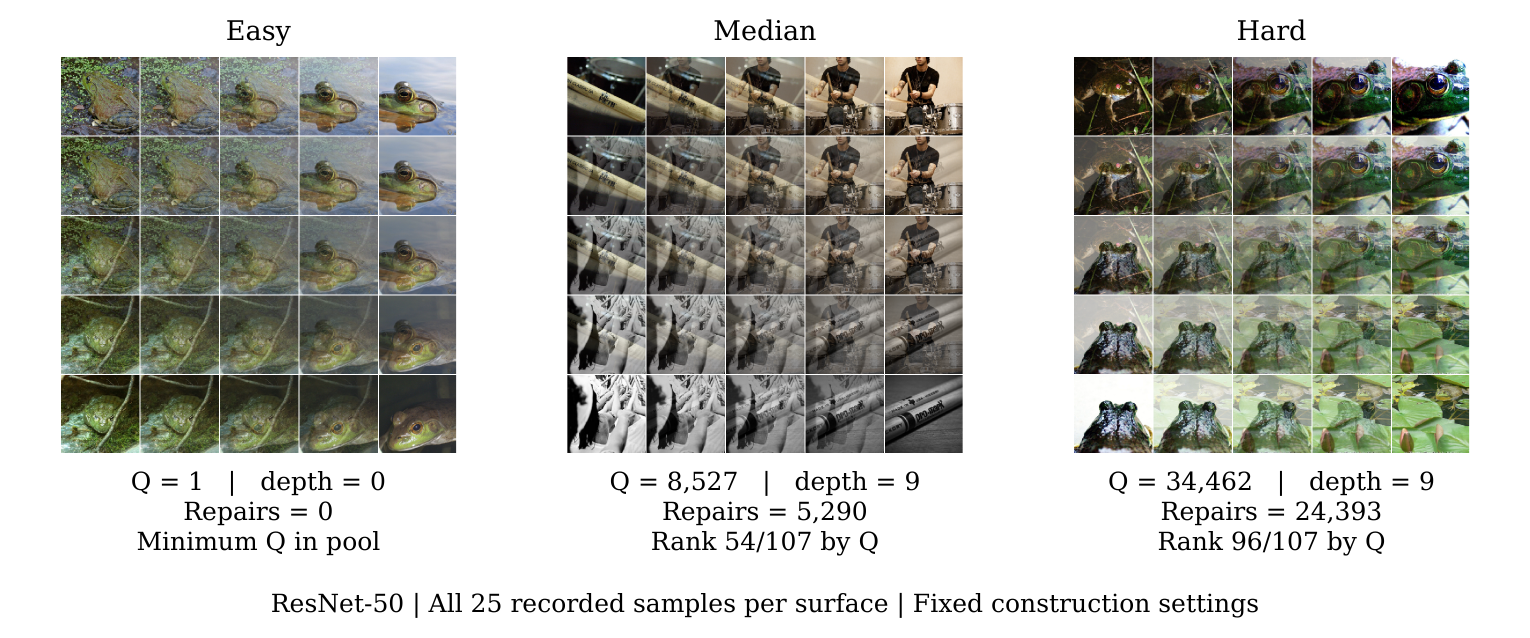}
\caption{Surface samples from easy, median and hard ResNet-50 constructions in the anchor-separation study. Each $5\times5$ grid contains $25$ on-label samples; panel labels report the predicted class, final quad count $Q$, refinement depth and repair attempts. The median example ranks $54/107$ among the available exports, and the hard example ranks $96/107$. These examples are separate from the $6000$-configuration ImageNet collection. Agreement at the displayed samples does not verify label preservation between them.}
\label{fig:animals}
\end{figure*}

\subsection{Trained versus random initialisation}
Loops in the trained networks are markedly easier to fill than in the mean-score-adjusted randomly initialised baselines under the same construction budget. Every one of the $2000$ trained-network loops is accepted at the fixed budget, against $877$ of $1520$ for the adjusted randomly initialised networks (Table~\ref{tab:cifar}); the remaining $643$ reach the parameter ceiling and are unresolved. An unresolved outcome does not show that a filling is impossible, but neither can it be counted as a success merely because the implementation returned a mesh.

\begin{table*}[t]
\centering
\small
\caption{CIFAR-100 construction outcomes at refinement depth $14$. Medians include all attempts, including unresolved constructions. Initial fraction denotes the sampled on-label fraction of the initial bilinear patch. Randomly initialised classifiers use the fixed mean-score adjustment.}
\label{tab:cifar}
\begin{tabular}{lrrrrr}
\toprule
Arm / seed & Tested & Accepted & Median quads & Median repairs & Initial fraction \\
\midrule
Trained / 0 & 1000 & 1000 & 7960.0 & 5238.5 & 0.8361 \\
Trained / 1 & 1000 & 1000 & 8482.0 & 5534.5 & 0.8127 \\
Adjusted random / 0 & 770 & 394 & 64564.0 & 54820.0 & 0.0579 \\
Adjusted random / 1 & 750 & 483 & 60689.5 & 50218.0 & 0.0735 \\
\bottomrule
\end{tabular}
\end{table*}

Pooled across seeds, adjusted random constructions require about seven and a half times as many quads and close to ten times as many vertex repairs, and their initial bilinear patch starts out $0.07$ on-label against $0.82$ for trained classifiers. These summaries include every recorded attempt, unresolved meshes among them, so they are not costs conditional on acceptance; the on-label figure is a fraction of samples within a patch, not a frequency of wholly on-label patches.

This comparison separates readily accepted trained-network constructions from more expensive and frequently unresolved constructions for the adjusted random classifiers. It establishes an operational difference in this dataset and architecture, subject to the score-adjustment and eligibility caveats of Section~\ref{sec:setup} and to the dependence of the measured costs on the construction algorithm and its budget.

\subsection{Checkpoints and within-class variation}
\begin{table*}[t]
\centering
\small
\caption{Construction effort across training checkpoints, using one configuration per eligible class for each seed and epoch. Root fraction denotes acceptance without subdivision. This sampling protocol differs from the ten-loop-per-class comparison; the archive does not support strict acceptance rates.}
\label{tab:checkpoints}
\begin{tabular}{lrrrrrr}
\toprule
Seed & Epoch & Tested & Accuracy & Root fraction & Median quads & Median repairs \\
\midrule
0 & 1 & 90 & 0.2522 & 0.156 & 25942.0 & 20826.0 \\
0 & 3 & 100 & 0.4196 & 0.150 & 6482.5 & 4242.5 \\
0 & 10 & 100 & 0.5496 & 0.040 & 15104.5 & 9913.0 \\
0 & 30 & 100 & 0.6779 & 0.040 & 15986.5 & 10244.5 \\
1 & 1 & 92 & 0.2455 & 0.130 & 23498.5 & 15975.0 \\
1 & 3 & 99 & 0.4309 & 0.192 & 16180.0 & 11344.0 \\
1 & 10 & 100 & 0.5729 & 0.170 & 9719.5 & 6256.5 \\
1 & 30 & 100 & 0.6829 & 0.040 & 18082.0 & 11318.0 \\
\bottomrule
\end{tabular}
\end{table*}
The checkpoint study does not identify an epoch at which fillability emerges. We analyse one configuration per eligible class at epochs $1$, $3$, $10$ and $30$ for both seeds (Table~\ref{tab:checkpoints}). Mesh size follows no monotone trajectory, and across the eight checkpoint summaries its Spearman correlation with test accuracy is $-0.31$, with root acceptance at $-0.41$. These summaries are few and mutually dependent, so they establish neither a predictor of accuracy nor an epoch at which fillability emerges. The checkpoint archive reports construction outcomes without the per-loop ceiling field needed to reproduce strict acceptance rates, so we make no retrospective strict-acceptance claim for that archive.

Variation within a single class is large. In the CIFAR endpoint study the median class spans a factor of roughly thirty between its most and least expensive loop for one seed and forty for the other, and about a third of classes disagree across their ten loops on whether subdivision is needed at all. The same holds at ImageNet scale: all $800$ additional cyclic-order constructions are accepted, yet merely reordering the same four anchors flips root acceptance in about one class in five, for both ConvNeXt-Tiny and Swin-T. A class therefore has no single characteristic cost, and an aggregate statistic describes a sampling protocol as much as it describes a region. Appendix~\ref{app:orderings} gives the per-ordering summaries and the variance decomposition.

\subsection{Threshold sensitivity and dense audits}
Finer resolution increases construction effort. Tightening the grey-RMS threshold $\tau$ from $1.0$ to $0.125$ raises the mean quad count of a fixed subset of $112$ ImageNet configurations by a factor of about thirty-six, and the deepest level used from $8$ to $11$. This paired subset contains configurations accepted at all four thresholds and measures cost sensitivity. Appendix~\ref{app:thresholds} reports the full attempt counts, including repair failures at each non-default threshold, for assessing construction outcomes separately from this cost comparison.

Dense resampling shows high observed label agreement in the two audited Swin-T configurations. On a $65\times65$ parameter grid for each construction, $8444$ of $8450$ samples ($99.93\%$) carry the target label. The six off-label samples all lie in patches accepted by size, whose grey-RMS diameter was at most $\tau$, locating the observed departures from label preservation within these sub-resolution patches (Appendix~\ref{app:visual}).

\subsection{A known-topology control}
Acceptance is informative only if the procedure can fail when a filling does not exist. We therefore run the same construction on an analytic region whose topology is known: an annulus in two coordinates crossed with $\mathbb{R}^{3070}$, carrying one loop that winds around the hole and one sector loop that does not. At $\tau=0.5$ and hole radius $12$ grey-RMS, the winding loop reaches the depth-$14$ ceiling with $14732$ patches accepted only there, while the sector loop is accepted at the root with a single quad. The implementation's legacy success flag is true in both cases, which is the concrete reason the parameter ceiling must be separated from geometric size acceptance.

In the recorded radius sweep, holes of radius $\tau$ or more leave the winding construction unresolved, whereas holes of radius $\tau/2$ or less are missed and the loop is accepted. This boundary is specific to the control geometry, loop and chosen threshold. It is not a universal minimum detectable feature size, and the missed range does not lie below an $8$-bit quantisation floor, because quantisation distance in this metric depends on dimension and direction (Appendix~\ref{app:controls}).

\section{Interpretation and limitations}
\label{sec:discussion}
The primary finding is that all sampled configurations across the pretrained classifiers yield accepted constructions under the finite-resolution criterion. This extends the empirical study of same-label paths \citep{fawzi2018empirical} to the question of whether loops can be filled within a decision region. Boundary geometry, path connectivity and loop fillability describe different aspects of classifier structure. A boundary can pass close to natural images while the region remains connected, and path connectivity alone does not exclude a hole around which a loop winds (Appendix~\ref{app:theory}). The present experiments therefore complement local adversarial analyses \citep{szegedy2013intriguing,moosavi2016deepfool} by examining two-dimensional constructions in input space.

Construction effort distinguishes configurations even when their acceptance outcomes agree. The trained and adjusted randomly initialised classifiers differ substantially in mesh size, repair count and unresolved outcomes under the same budget. Within a class, changing the anchors or their cyclic ordering also changes the work required. These costs depend on the classifier, anchor-selection protocol and construction algorithm; they do not define an intrinsic complexity of the region. Score adjustment and eligible anchors also differ in the trained-versus-random comparison, limiting attribution to training alone.

Sampling also shapes the interpretation of aggregate costs. Deterministic selection gives reproducible configurations, while the repeated-loop and distance-stratified studies expose variation within classes. Representation separation is associated with boundary proximity in the selected samples, leaving their contributions to cost coupled (Appendix~\ref{app:distance}). The geometric diagnostics provide a complementary comparison with the Coons reference: the ImageNet curvature results show little evidence of a systematic bending-energy difference in the selected pool, while the area ratio is affected by discretisation (Appendix~\ref{app:curvature}). Construction effort and these geometric summaries therefore answer different questions; the latter do not establish an intrinsic complexity or generalisation measure.

A continuous label-preserving filling would connect alternative same-label interpolation paths through a deformation with fixed endpoints. In our construction the boundary loop is fixed before surface filling, while interior vertices may be displaced by repair to restore the target prediction. The computed patches therefore provide finite-resolution candidates for a disk spanning that prescribed loop, as well as a way to inspect two-parameter families of interpolated images. Applications to robustness, adversarial attacks or semantic image generation would require separate evaluation.

The analytic annulus supplies a known obstruction: in the recorded sweep, winding configurations are unresolved at the tested hole radii of $\tau$ and above, while sector configurations are accepted. Its radius sweep measures sensitivity for that control geometry, rather than a universal detection bound for learned decision regions. The threshold study measures the cost of finer resolution, while dense resampling and edge-mismatch audits assess label agreement and continuity in the returned meshes. Certifying a continuous filling would require label preservation between samples and a conforming mesh across unequal refinement levels.

Further experiments can separate factors that are combined here. CIFAR-100 and ImageNet differ in input dimension by about a factor of $49$, from $3072$ to $150528$, but also differ in architecture, label set and sampling protocol. Classifiers trained to matched accuracy across resolutions would make the role of input dimension more directly testable. Extending the construction to structured prediction would instead require defining regions through an exact semantic mask or a coarser layout, with a repair objective that preserves that output.

\section{Conclusion}
\label{sec:conclusion}
We extend the empirical study of path connectivity in image-classifier decision regions to loop fillability. Across $6000$ configurations from six pretrained ImageNet classifiers, every construction satisfies the finite-resolution acceptance criterion. The analytic annulus control leaves winding loops unresolved at the tested hole radii at or above the resolution threshold. Together, these results provide empirical evidence consistent with simply connected decision regions at the tested resolution, while distinguishing accepted constructions from termination at the refinement ceiling.

Construction effort reveals differences that acceptance alone does not capture. Under the same budget, all $2000$ trained CIFAR-100 configurations are accepted, compared with $877$ of $1520$ configurations from mean-score-adjusted randomly initialised classifiers. The pooled median quad and repair counts for the adjusted random classifiers are about $7.5$ and $9.7$ times those of the trained classifiers, respectively. Effort also varies substantially within classes and with the cyclic ordering of fixed anchors. These measurements characterise the tested classifiers and sampling protocols; they do not isolate the causal effect of training or define intrinsic region complexity.

Dense resampling and continuity audits identify the remaining requirements for a continuous, label-preserving disk extension: verified labels between samples and a conforming mesh. Matched-accuracy studies across input resolutions would help separate dimension from classifier differences. The present study provides a constructive approach to investigating loop fillability and measuring its cost, extending empirical analysis of decision regions beyond whether same-label inputs can be joined by a path.

\section*{Code availability}
Our code is available at the following URL: \url{https://github.com/mdppml/contractible-class-regions}. The experimental protocols, classifier adjustment and control definitions are described in the appendices.

\label{page:main-end}
\clearpage
\section*{Acknowledgments}
This research was supported by the German Federal Ministry of Education and Research (BMBF) under project number 01ZZ2010. Arjhun Swaminathan additionally acknowledges financial support from Coefficient Giving during the preparation of this version of the manuscript. The authors acknowledge use of the Training Center for Machine Learning (TCML) cluster at the University of T\"ubingen.

\section*{AI Use Statement}
Generative AI tools were used only for grammar, clarity, and stylistic editing of the manuscript. The authors reviewed all suggested changes and take full responsibility for the final text and scientific content.

\bibliographystyle{plainnat}
\bibliography{references}

@incollection{barnhill1977representation,
  title={Representation and approximation of surfaces},
  author={Barnhill, Robert E},
  booktitle={Mathematical software},
  pages={69--120},
  year={1977},
  publisher={Elsevier}
}

@article{beise2021decision,
  title={On decision regions of narrow deep neural networks},
  author={Beise, Hans-Peter and Da Cruz, Steve Dias and Schr{\"o}der, Udo},
  journal={Neural Networks},
  volume={140},
  pages={121--129},
  year={2021},
  publisher={Elsevier}
}

@article{bianchini2014complexity,
  title={On the complexity of neural network classifiers: A comparison between shallow and deep architectures},
  author={Bianchini, Monica and Scarselli, Franco},
  journal={IEEE transactions on neural networks and learning systems},
  volume={25},
  number={8},
  pages={1553--1565},
  year={2014},
  publisher={IEEE}
}

@article{brendel2017decision,
  title={Decision-based adversarial attacks: Reliable attacks against black-box machine learning models},
  author={Brendel, Wieland and Rauber, Jonas and Bethge, Matthias},
  journal={arXiv preprint arXiv:1712.04248},
  year={2017}
}

@article{carlsson2009topology,
  title={Topology and data},
  author={Carlsson, Gunnar},
  journal={Bulletin of the American Mathematical Society},
  volume={46},
  number={2},
  pages={255--308},
  year={2009}
}

@inproceedings{chen2020hopskipjumpattack,
  title={{HopSkipJumpAttack}: A query-efficient decision-based attack},
  author={Chen, Jianbo and Jordan, Michael I and Wainwright, Martin J},
  booktitle={2020 ieee symposium on security and privacy (sp)},
  pages={1277--1294},
  year={2020},
  organization={IEEE}
}

@techreport{coons1967surfaces,
  title={Surfaces for Computer-Aided Design of Space Forms},
  author={Coons, Steven A.},
  institution={Massachusetts Institute of Technology, Project MAC},
  number={MAC-TR-041},
  year={1967},
  url={https://hdl.handle.net/1721.1/149362}
}

@inproceedings{deng2009imagenet,
  title={{ImageNet}: A large-scale hierarchical image database},
  author={Deng, Jia and Dong, Wei and Socher, Richard and Li, Li-Jia and Li, Kai and Fei-Fei, Li},
  booktitle={2009 IEEE conference on computer vision and pattern recognition},
  pages={248--255},
  year={2009},
  organization={IEEE}
}

@article{dosovitskiy2020image,
  title={An image is worth 16x16 words: Transformers for image recognition at scale},
  author={Dosovitskiy, Alexey and Beyer, Lucas and Kolesnikov, Alexander and Weissenborn, Dirk and Zhai, Xiaohua and Unterthiner, Thomas and Dehghani, Mostafa and Minderer, Matthias and Heigold, Georg and Gelly, Sylvain and others},
  journal={arXiv preprint arXiv:2010.11929},
  year={2020}
}

@book{edelsbrunner2010computational,
  title={Computational topology: an introduction},
  author={Edelsbrunner, Herbert and Harer, John},
  year={2010},
  publisher={American Mathematical Soc.}
}

@inproceedings{fawzi2018empirical,
  title={Empirical study of the topology and geometry of deep networks},
  author={Fawzi, Alhussein and Moosavi-Dezfooli, Seyed-Mohsen and Frossard, Pascal and Soatto, Stefano},
  booktitle={Proceedings of the IEEE Conference on Computer Vision and Pattern Recognition},
  pages={3762--3770},
  year={2018}
}

@article{ghrist2008barcodes,
  title={Barcodes: the persistent topology of data},
  author={Ghrist, Robert},
  journal={Bulletin of the American Mathematical Society},
  volume={45},
  number={1},
  pages={61--75},
  year={2008}
}

@article{goodfellow2014explaining,
  title={Explaining and harnessing adversarial examples},
  author={Goodfellow, Ian J and Shlens, Jonathon and Szegedy, Christian},
  journal={arXiv preprint arXiv:1412.6572},
  year={2014}
}

@book{hatcher2005algebraic,
  title={Algebraic topology},
  author={Hatcher, Allen},
  year={2002},
  publisher={Cambridge University Press}
}

@inproceedings{he2016deep,
  title={Deep residual learning for image recognition},
  author={He, Kaiming and Zhang, Xiangyu and Ren, Shaoqing and Sun, Jian},
  booktitle={Proceedings of the IEEE conference on computer vision and pattern recognition},
  pages={770--778},
  year={2016}
}

@inproceedings{huang2017densely,
  title={Densely connected convolutional networks},
  author={Huang, Gao and Liu, Zhuang and Van Der Maaten, Laurens and Weinberger, Kilian Q},
  booktitle={Proceedings of the IEEE conference on computer vision and pattern recognition},
  pages={4700--4708},
  year={2017}
}

@techreport{krizhevsky2009learning,
  author={Krizhevsky, Alex},
  title={Learning Multiple Layers of Features from Tiny Images},
  institution={University of Toronto},
  year={2009},
  url={https://www.cs.toronto.edu/~kriz/learning-features-2009-TR.pdf}
}

@article{krizhevsky2012imagenet,
  title={{ImageNet} classification with deep convolutional neural networks},
  author={Krizhevsky, Alex and Sutskever, Ilya and Hinton, Geoffrey E},
  journal={Advances in neural information processing systems},
  volume={25},
  year={2012}
}

@inproceedings{liu2021swin,
  title={{Swin Transformer}: Hierarchical vision transformer using shifted windows},
  author={Liu, Ze and Lin, Yutong and Cao, Yue and Hu, Han and Wei, Yixuan and Zhang, Zheng and Lin, Stephen and Guo, Baining},
  booktitle={Proceedings of the IEEE/CVF international conference on computer vision},
  pages={10012--10022},
  year={2021}
}

@inproceedings{liu2022convnet,
  title={A {ConvNet} for the 2020s},
  author={Liu, Zhuang and Mao, Hanzi and Wu, Chao-Yuan and Feichtenhofer, Christoph and Darrell, Trevor and Xie, Saining},
  booktitle={Proceedings of the IEEE/CVF conference on computer vision and pattern recognition},
  pages={11976--11986},
  year={2022}
}

@inproceedings{maho2021surfree,
  title={{SurFree}: a fast surrogate-free black-box attack},
  author={Maho, Thibault and Furon, Teddy and Le Merrer, Erwan},
  booktitle={Proceedings of the IEEE/CVF conference on computer vision and pattern recognition},
  pages={10430--10439},
  year={2021}
}

@inproceedings{moosavi2016deepfool,
  title={{DeepFool}: a simple and accurate method to fool deep neural networks},
  author={Moosavi-Dezfooli, Seyed-Mohsen and Fawzi, Alhussein and Frossard, Pascal},
  booktitle={Proceedings of the IEEE conference on computer vision and pattern recognition},
  pages={2574--2582},
  year={2016}
}

@inproceedings{moosavi2017universal,
  title={Universal adversarial perturbations},
  author={Moosavi-Dezfooli, Seyed-Mohsen and Fawzi, Alhussein and Fawzi, Omar and Frossard, Pascal},
  booktitle={Proceedings of the IEEE conference on computer vision and pattern recognition},
  pages={1765--1773},
  year={2017}
}

@inproceedings{nguyen2018neural,
  title={Neural networks should be wide enough to learn disconnected decision regions},
  author={Nguyen, Quynh and Mukkamala, Mahesh Chandra and Hein, Matthias},
  booktitle={International conference on machine learning},
  pages={3740--3749},
  year={2018},
  organization={PMLR}
}

@inproceedings{rahmati2020geoda,
  title={{GeoDA}: a geometric framework for black-box adversarial attacks},
  author={Rahmati, Ali and Moosavi-Dezfooli, Seyed-Mohsen and Frossard, Pascal and Dai, Huaiyu},
  booktitle={Proceedings of the IEEE/CVF conference on computer vision and pattern recognition},
  pages={8446--8455},
  year={2020}
}

@inproceedings{ramamurthy2019topological,
  title={Topological data analysis of decision boundaries with application to model selection},
  author={Ramamurthy, Karthikeyan Natesan and Varshney, Kush and Mody, Krishnan},
  booktitle={International Conference on Machine Learning},
  pages={5351--5360},
  year={2019},
  organization={PMLR}
}

@inproceedings{reza2023cgba,
  title={CGBA: Curvature-aware geometric black-box attack},
  author={Reza, Md Farhamdur and Rahmati, Ali and Wu, Tianfu and Dai, Huaiyu},
  booktitle={Proceedings of the IEEE/CVF international conference on computer vision},
  pages={124--133},
  year={2023}
}

@article{russakovsky2015imagenet,
  title={{ImageNet} large scale visual recognition challenge},
  author={Russakovsky, Olga and Deng, Jia and Su, Hao and Krause, Jonathan and Satheesh, Sanjeev and Ma, Sean and Huang, Zhiheng and Karpathy, Andrej and Khosla, Aditya and Bernstein, Michael and others},
  journal={International journal of computer vision},
  volume={115},
  number={3},
  pages={211--252},
  year={2015},
  publisher={Springer}
}

@article{simonyan2014very,
  title={Very deep convolutional networks for large-scale image recognition},
  author={Simonyan, Karen and Zisserman, Andrew},
  journal={arXiv preprint arXiv:1409.1556},
  year={2014}
}

@article{swaminathan2025accelerating,
  title={Accelerating Targeted Hard-Label Adversarial Attacks in Low-Query Black-Box Settings},
  author={Swaminathan, Arjhun and Akg{\"u}n, Mete},
  journal={arXiv preprint arXiv:2505.16313},
  year={2025}
}

@InProceedings{swaminathan2025dynamic,
author="Swaminathan, Arjhun
and Akg{\"u}n, Mete",
title="Dynamic k-Anonymity for Electronic Health Records: A Topological Framework",
booktitle="Computer Security. ESORICS 2024 International Workshops",
year="2025",
publisher="Springer Nature Switzerland",
address="Cham",
pages="137--152",
isbn="978-3-031-82349-7"
}

@article{szegedy2013intriguing,
  title={Intriguing properties of neural networks},
  author={Szegedy, Christian and Zaremba, Wojciech and Sutskever, Ilya and Bruna, Joan and Erhan, Dumitru and Goodfellow, Ian and Fergus, Rob},
  journal={arXiv preprint arXiv:1312.6199},
  year={2013}
}

@inproceedings{szegedy2015going,
  title={Going deeper with convolutions},
  author={Szegedy, Christian and Liu, Wei and Jia, Yangqing and Sermanet, Pierre and Reed, Scott and Anguelov, Dragomir and Erhan, Dumitru and Vanhoucke, Vincent and Rabinovich, Andrew},
  booktitle={Proceedings of the IEEE conference on computer vision and pattern recognition},
  pages={1--9},
  year={2015}
}

@inproceedings{tan2019efficientnet,
  title={{EfficientNet}: Rethinking model scaling for convolutional neural networks},
  author={Tan, Mingxing and Le, Quoc},
  booktitle={International conference on machine learning},
  pages={6105--6114},
  year={2019},
  organization={PMLR}
}

@article{watanabe2022topological,
  title={Topological measurement of deep neural networks using persistent homology},
  author={Watanabe, Satoru and Yamana, Hayato},
  journal={Annals of Mathematics and Artificial Intelligence},
  volume={90},
  number={1},
  pages={75--92},
  year={2022},
  publisher={Springer}
}

\clearpage
\appendix
\onecolumn
\aistatstitle{Empirical Evidence for Simply Connected Decision Regions in Image Classifiers: \ Supplementary Materials}
\section{Why a two-dimensional probe remains relevant}
\label{app:theory}
A loop is null-homotopic when its map from the circle extends continuously to the disk. This is the exact property motivating a spanning surface; it is not a claim that two dimensions approximate the ambient geometry \citep{hatcher2005algebraic}. For $D\geq2$, the complement of a codimension-two affine subspace is homeomorphic to $(\mathbb{R}^2\setminus\{0\})\times\mathbb{R}^{D-2}$. Radial projection in the first factor, together with contraction of the second, gives a deformation retraction onto a circle. The complement is therefore path connected and contains non-contractible loops. High ambient dimension alone does not remove the distinction between path connectivity and loop fillability. The finite computation in the main paper does not establish a disk extension for every tested boundary.

\section{Sampling protocols and random-initialisation diagnostics}
\label{app:sampling}
\paragraph{CIFAR-100 architecture and training.}
The archived training driver uses the standard $50000$ training and $10000$ test images. Its VGG-style network has three blocks with $64$, $128$ and $256$ channels. Each block contains two $3\times3$ convolutions with padding $1$ and no bias, each followed by batch normalisation and ReLU, then $2\times2$ max pooling. Global average pooling and a $256$-to-$100$ linear layer produce the logits. The driver specifies $30$ epochs of cross-entropy training with Nesterov SGD, initial learning rate $0.05$, momentum $0.9$, weight decay $5\times10^{-4}$, batch size $128$ and cosine learning-rate decay over the $30$ epochs. The training order is shuffled each epoch; the only augmentation is a horizontal flip applied to an entire batch with probability $0.5$. Channel means are $(0.5071,0.4865,0.4409)$ and standard deviations $(0.2673,0.2564,0.2762)$. Seeds are $0$ and $1$, and checkpoints are saved at epochs $0$, $1$, $3$, $10$ and $30$. The compact architecture and short schedule were chosen for the training comparison rather than peak benchmark accuracy. The supplementary training driver records these settings.

ImageNet selection is model specific. For each target label we take the first four images the model predicts as that label, keeping only those whose prediction is unchanged after being saved and reloaded, subject to the substitutions in Appendix~\ref{app:provenance}. It is not based on a uniform draw from the class region, confidence matching, or an intrinsic difficulty score.

For the CIFAR endpoint study, six stable images form an anchor pool and ten distinct four-anchor cycles are enumerated from it. These configurations can share anchors and are not statistically independent samples of a region. The eligible-class counts are $100$, $100$, $77$ and $75$ for trained seed $0$, trained seed $1$, untrained seed $0$ and untrained seed $1$, respectively. All eligible construction attempts are retained in the denominators. Eligibility is determined before the construction outcome.

For random initialisation, write the logits as $g_k(x)$ and let $\mu_k$ be their fixed mean over the test pool. The adjusted hard-label classifier is $f_{\mathrm{adj}}(x)=\operatorname*{argmax}_k[g_k(x)-\mu_k]$. The same fixed vector is used for selection and construction. This adjustment defines different decision regions from the raw argmax classifier. The raw seed-$0$ CIFAR classifier predicts one class for all $10000$ test images and supplies no useful diversity of regions for the intended comparison. An additional raw random-weight ImageNet diagnostic likewise concentrates $97.7\%$ and $98.1\%$ of predictions in one class for two seeds and supplies only one tested loop per seed. Its trivial root acceptance is not comparable evidence for learned topology.

The checkpoint study uses one configuration per eligible class rather than the six-anchor, ten-cycle endpoint design. Both epoch-$1$ archives are present: $90$ configurations for seed $0$ and $92$ for seed $1$. Later counts are shown in Table~\ref{tab:checkpoints}. A change from the four-anchor to the six-anchor protocol changes the sampled initial-patch fractions; values from the two designs are not directly pooled.

\section{Acceptance controls and sensitivity to unresolved patches}
\label{app:controls}
The analytic control has target region $\{z\in\mathbb{R}^2:r_{\mathrm{in}}<\|z\|\leq r_{\mathrm{out}}\}\times\mathbb{R}^{3070}$, with a loop winding around its hole and a sector loop of winding zero. At hole radius $12$ grey-RMS and $\tau=0.5$, the winding case produces $73249$ leaf quads, including $14732$ accepted only at the depth-$14$ ceiling. It is unresolved under the main-paper criterion, although the implementation's legacy success flag is true. The sector case is accepted at the root with one quad. Thus, increasing repair strength or obtaining a returned mesh cannot by itself justify an accepted-filling claim.

\begin{table}[ht]
\centering
\small
\caption{Sensitivity to hole radius in the analytic annulus control at $\tau=0.5$. Outcomes use the finite-resolution acceptance criterion. The sector configuration is accepted without subdivision at every radius.}
\label{tab:controls}
\begin{tabularx}{\linewidth}{@{}>{\raggedright\arraybackslash}X>{\raggedright\arraybackslash}X>{\raggedright\arraybackslash}X>{\raggedright\arraybackslash}X@{}}
\toprule
Hole radius & Radius / $\tau$ & Winding outcome & Ceiling patches \\
\midrule
12 & 24 & Unresolved & 14732 \\
4 & 8 & Unresolved & 1456 \\
1 & 2 & Unresolved & 96 \\
0.5 & 1 & Unresolved & 24 \\
0.25 & 0.5 & Accepted; hole missed & 0 \\
0.1 & 0.2 & Accepted; hole missed & 0 \\
0.05 & 0.1 & Accepted; hole missed & 0 \\
\bottomrule
\end{tabularx}
\end{table}

Table~\ref{tab:controls} reports the radius sweep, which is specific to this analytic region and loop geometry. It is not a universal detection bound: missing a small hole in this example does not establish a minimum topological feature size for other classifiers. Nor does a sub-grey-RMS radius fall below a universal $8$-bit quantisation floor. In $3072$ raw pixel coordinates, changing one coordinate by one grey level has unweighted RMS distance $1/\sqrt{3072}$, far below half a grey level. Quantisation distances depend on dimension and direction, whereas the construction operates on continuous interpolations.

The projected-network diagnostic confines the decision to two coordinates, unlike a full-image trained approximation that may bridge the hole off-plane. Its finite planar audit checks sampled circle labels and off-label interior points; it does not certify labels between samples. The analytic classifier supplies the known topology by construction. The supplementary files include \texttt{supplementary/holefill.py}, the analytic and projected-network definitions, archived numeric outcomes and reproduction settings. Results for alternative repair or boundary-construction settings constitute separate comparisons.

\section{Multiple orderings of the same ImageNet anchors}
\label{app:orderings}
Four fixed anchors admit three distinct undirected four-cycles. We compare the archived ordering with both alternatives for $200$ classes in each of ConvNeXt-Tiny and Swin-T. The study contains $1200$ configurations in total: $400$ archived and $800$ additional constructions. All additional constructions satisfy the acceptance criterion. Table~\ref{tab:orderings} preserves the distinction between baseline and additional rows.

\begin{table}[ht]
\centering
\small
\caption{Construction outcomes for the three distinct undirected cyclic orderings of each fixed ImageNet anchor set. Ordering $0$ is the original ordering; orderings $1$ and $2$ add $800$ constructions to the $400$ original ones.}
\label{tab:orderings}
\begin{tabularx}{\linewidth}{@{}>{\raggedright\arraybackslash}X>{\raggedright\arraybackslash}X>{\raggedright\arraybackslash}X>{\raggedright\arraybackslash}X>{\raggedright\arraybackslash}X>{\raggedright\arraybackslash}X@{}}
\toprule
Model & Ordering & Tested & Root fraction & Median quads & Median repairs \\
\midrule
ConvNeXt-Tiny & 0 & 200 & 0.380 & 1001.5 & 689.5 \\
ConvNeXt-Tiny & 1 & 200 & 0.410 & 535.0 & 359.5 \\
ConvNeXt-Tiny & 2 & 200 & 0.380 & 836.5 & 556.0 \\
Swin-T & 0 & 200 & 0.415 & 395.5 & 241.5 \\
Swin-T & 1 & 200 & 0.385 & 359.5 & 225.5 \\
Swin-T & 2 & 200 & 0.415 & 514.0 & 326.5 \\
\bottomrule
\end{tabularx}
\end{table}

The median within-class largest-to-smallest quad-count ratio is $1.9$ for both models. Its $90$th percentile is $1321$ for ConvNeXt-Tiny and $607$ for Swin-T. Root acceptance differs across orderings in $45/200$ ConvNeXt-Tiny classes and $41/200$ Swin-T classes. For log quad count, $85.8\%$ of variance is between classes and $14.2\%$ within them for ConvNeXt-Tiny, compared with $88.3\%$ and $11.7\%$ for Swin-T. These fixed-anchor orderings do not measure variability over newly selected ImageNet anchor sets. The CIFAR six-anchor design has a different sampling distribution and gives $67\%$ between-class and $33\%$ within-class variance; the difference cannot be attributed to dataset alone.

\section{Anchor separation and its selection confound}
\label{app:distance}
The ResNet-50 stratified study initially selects $40$ candidate quadruples per group. The same save-and-reload stability check leaves $37$, $37$, $35$ and $26$ eligible configurations, totalling $135$. The first three groups vary minimum pairwise input-space separation; the fourth increases cosine separation in the penultimate representation while approximately matching input separation to the middle group. All $135$ recorded constructions satisfy the acceptance criterion, and Table~\ref{tab:distance} reports the four groups.

\begin{table}[ht]
\centering
\small
\caption{ResNet-50 construction effort under input- and representation-separation sampling. Separation and cost columns report medians; root fraction is the proportion accepted without subdivision.}
\label{tab:distance}
\begin{tabularx}{\linewidth}{@{}>{\raggedright\arraybackslash}X>{\raggedright\arraybackslash}X>{\raggedright\arraybackslash}X>{\raggedright\arraybackslash}X>{\raggedright\arraybackslash}X>{\raggedright\arraybackslash}X>{\raggedright\arraybackslash}X@{}}
\toprule
Selection & Tested & Input sep. & Repr. sep. & Root frac. & Median quads & Median repairs \\
\midrule
Low input & 37 & 59.10 & 0.317 & 0.216 & 1816 & 1189 \\
Middle input & 37 & 77.19 & 0.349 & 0.189 & 5404 & 3813 \\
High input & 35 & 114.71 & 0.389 & 0.257 & 4282 & 2882 \\
High representation & 26 & 74.33 & 0.644 & 0.038 & 12270 & 8844 \\
\bottomrule
\end{tabularx}
\end{table}

Input-space separation has Spearman correlations $0.029$ with quad count and $0.019$ with repairs in this sample. Median cost is larger in the high-representation group than in the middle input group, but their samples are small and the high-representation group is enriched for near-boundary anchors after filtering. The comparison therefore measures representation separation together with boundary proximity. It does not establish representation distance as an independent cause of construction effort. All groups use one architecture, and the three main-paper examples are drawn from this additional study.

\section{Visualisation, dense resampling and continuity audits}
\label{app:visual}
The main-paper examples retain all $25$ recorded surface samples, without changing image pixels or predictions. The easy, median and hard examples have final quad counts $Q=1$, $8527$ and $34462$, observed depths $0$, $9$ and $9$, and vertex-repair counts $0$, $5290$ and $24393$, respectively. The median is computed over the $107$ available ResNet-50 surface exports from the selected distance study, rather than over all ImageNet constructions. The hard example is a high-cost illustration (rank $96/107$), not the maximum-cost run. The easy and hard examples target the same bullfrog label with different anchor sets.

Separate audits resample Swin-T configurations $134$ and $437$ on a $65\times65$ parameter grid. They contain $0/4225$ and $6/4225$ off-label samples respectively, with the six misses lying within depth-$9$ sub-$\tau$ patches accepted by size. The displayed coarse grid for configuration $437$ also contains one off-label sample ($24/25$ agreement). These observations remain part of the record even though the three selected main-paper examples have $25/25$ agreement. Table~\ref{tab:visual} reports the mesh diagnostics for the three reconstructed configurations.

\begin{table}[ht]
\centering
\small
\caption{Mesh diagnostics for three reconstructed configurations. Displacement is measured from parent average to final vertex. Crack size measures the mismatch along edges shared by patches at unequal refinement levels.}
\label{tab:visual}
\begin{tabularx}{\linewidth}{@{}>{\raggedright\arraybackslash}X>{\raggedright\arraybackslash}X>{\raggedright\arraybackslash}X>{\raggedright\arraybackslash}X>{\raggedright\arraybackslash}X@{}}
\toprule
Configuration & Final quads & Repairs & Max displacement / $\tau$ & Max crack / $\tau$ \\
\midrule
Swin-T 134 & 364 & 199 & 0.0099 & 0 \\
Swin-T 437 & 14161 & 10579 & 0.0303 & $1.1\times10^{-6}$ \\
DenseNet-121 934 & 14680 & 11292 & 0.0957 & 0 \\
\bottomrule
\end{tabularx}
\end{table}

\begin{figure}[!htbp]
\centering
\includegraphics[width=0.88\linewidth]{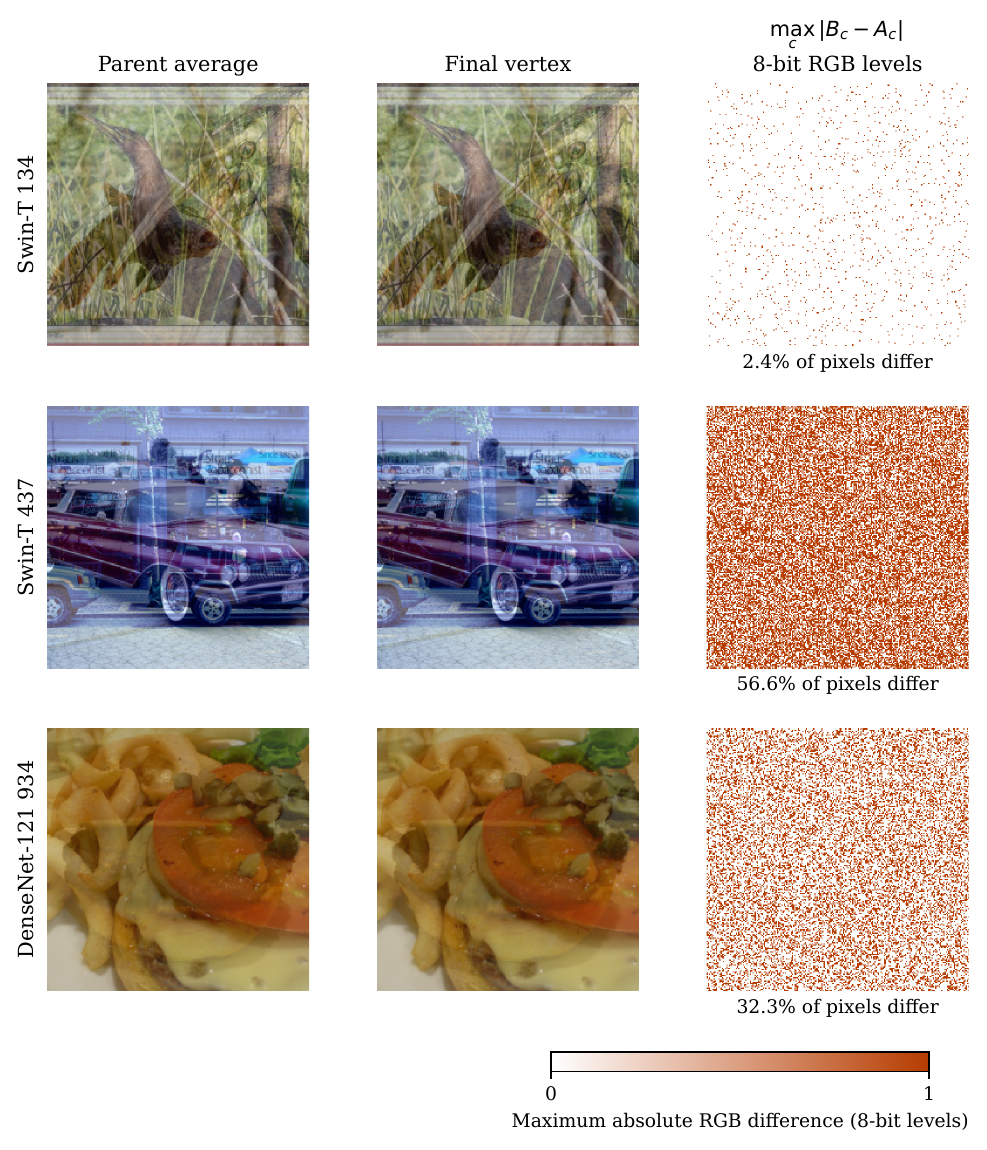}
\caption{Reconstructed parent averages $A$ (left), final vertices $B$ (middle) and their display differences (right) for Swin-T $134$, Swin-T $437$ and DenseNet-121 $934$. Each difference panel shows the maximum absolute RGB-channel change per pixel, on a common scale from zero to one $8$-bit level; percentages give the fraction of pixels that differ. Parent averages need not coincide with the initial boundary-repair states, and differences between saved renderings do not measure full-precision displacements.}
\label{fig:repairs}
\end{figure}

The displacement diagnostic reconstructs a newly introduced vertex's parent average and compares it with its final stored value. For repaired interior vertices this measures the local displacement induced by repair. The largest reconstructed displacement in these audited cases is $0.0957\tau$. This is a local mesh diagnostic, not a bound on every repair or on the total deformation of the surface. Figure~\ref{fig:repairs} shows one recorded parent-average/final-vertex pair from each audited configuration, alongside colour-coded differences of the saved $8$-bit renderings. At each pixel, the diagnostic reports the maximum absolute difference across RGB channels. All three panels share a scale from zero (white) to one $8$-bit level (orange), and report the percentage of pixels that differ. These display differences are affected by quantisation and do not measure the full-precision displacement.

\begin{samepage}
At unequal refinement levels, the maximum audited edge mismatch is zero for Swin-T $134$ and DenseNet-121 $934$, and $1.1\times10^{-6}\tau$ for Swin-T $437$. These small observations do not prove structural continuity in other meshes. Near-boundary floating-point decisions also make large reconstructed meshes non-bit-reproducible. This is separate from the six off-label interior samples, which arise in size-accepted patches rather than stored-vertex prediction changes.
\par\end{samepage}

\FloatBarrier
\section{Area and curvature diagnostics}
\label{app:curvature}
We do not use the constructed-to-Coons area ratio as a geometric-closeness statistic. The earlier comparison used different discretisations for the adaptive construction and the reference. In the diagnostic archive, the ratio has Spearman correlation $-0.868$ with quad count and an upper mode explained by a high-dimensional discretisation effect, with reference limit $1.3523308$. Similar area does not imply similar geometry. Let $\rho$ denote the constructed-to-Coons area ratio. Table~\ref{tab:area_diagnostic} reports the per-model values; between-model differences in this statistic do not establish model complexity.

\begin{table}[ht]
\centering
\small
\caption{Constructed-to-Coons area ratios across architectures, using the $5939$ original successful runs and excluding the $61$ refill records. The adaptive and reference surfaces use different discretisations; these ratios do not measure geometric closeness or intrinsic complexity.}
\label{tab:area_diagnostic}
\begin{tabularx}{\linewidth}{@{}>{\raggedright\arraybackslash}X>{\raggedright\arraybackslash}X>{\raggedright\arraybackslash}X>{\raggedright\arraybackslash}X>{\raggedright\arraybackslash}X@{}}
\toprule
Model & Archive $n$ & Spearman $\rho,Q$ & Median $\rho$ & Median quads \\
\midrule
ResNet-50 & 999 & -0.817 & 1.0154 & 8101.0 \\
DenseNet-121 & 975 & -0.799 & 1.0102 & 15430.0 \\
EfficientNet-B0 & 990 & -0.764 & 1.0115 & 14764.0 \\
ConvNeXt-Tiny & 992 & -0.865 & 1.0657 & 980.5 \\
ViT-B/16 & 993 & -0.826 & 1.0447 & 2260.0 \\
Swin-T & 990 & -0.881 & 1.0710 & 359.5 \\
\bottomrule
\end{tabularx}
\end{table}

We instead evaluate curvature diagnostics using the same triangulation for a construction and its Coons reference. The discrete Willmore diagnostic integrates squared mean-curvature-vector magnitude and is scale invariant under uniform spatial dilation; area-normalised bending energy does not have this property. For the available $110$-configuration ImageNet curvature pool, $107$ configurations can be reconstructed and three lack archived anchors. This is a dedicated pool, not every subdivided configuration in the $6000$-configuration study. The median per-configuration Willmore ratio is $1.0000$; $62/107$ lie within $1\%$ of unity and the paired signed-rank comparison gives $p=0.302$. The ratio of total energies is $0.948$, which is a different summary from the median of ratios.

The CIFAR archive contains $20$ configurations from each of five checkpoint arms: seed $0$ at epochs $3$, $10$ and $30$, and seed $1$ at epochs $0$ and $30$. The epoch-$0$ arm uses the raw classifier and all its configurations target label $94$; it is distinct from the adjusted random classifiers in Table~\ref{tab:cifar}. Across all $100$ configurations, the Willmore ratio has median $0.275$ and ranges from $0.031$ to $1.569$. Arm medians range from $0.130$ to $0.906$. Table~\ref{tab:curvature} also reports ratios of mean absolute interior angle deficit, a discrete integrated-Gaussian-curvature quantity, and median mean-curvature-vector magnitude. The angle deficits are not divided by vertex area and should not be interpreted as pointwise curvature densities. These are supplementary diagnostics rather than proposed predictors of generalisation. A between-dataset contrast may reflect classifier, sampling, normalisation, discretisation and dimension differences; it does not isolate dimension. The pooled CIFAR diagnostic likewise does not isolate training or supply an intrinsic complexity measure.

\begin{table}[ht]
\centering
\small
\caption{Curvature ratios between constructed surfaces and their Coons references on matching $65\times65$ triangulations. The angle-deficit and mean-curvature columns report median per-configuration ratios of the indicated spatial summaries. The Willmore columns report the median and range of per-configuration energy ratios.}
\label{tab:curvature}
\begin{tabularx}{\linewidth}{@{}>{\raggedright\arraybackslash}X>{\raggedright\arraybackslash}X>{\raggedright\arraybackslash}X>{\raggedright\arraybackslash}X>{\raggedright\arraybackslash}X>{\raggedright\arraybackslash}X>{\raggedright\arraybackslash}X@{}}
\toprule
Pool / checkpoint & Tested & Mean $|\delta|$ ratio & Median $|H|$ ratio & Median $W$ ratio & Min $W$ ratio & Max $W$ ratio \\
\midrule
ImageNet selected pool & 107 & 1.0000 & 1.0000 & 1.0000 & 0.3859 & 1.0628 \\
CIFAR s0 / epoch 3 & 20 & 1.0007 & 0.9141 & 0.9062 & 0.0500 & 1.1304 \\
CIFAR s0 / epoch 10 & 20 & 1.0034 & 0.6936 & 0.3829 & 0.0311 & 1.0947 \\
CIFAR s0 / epoch 30 & 20 & 1.0113 & 0.6308 & 0.2689 & 0.0358 & 1.0004 \\
CIFAR s1 / epoch 0 & 20 & 1.1971 & 0.4781 & 0.1296 & 0.0677 & 1.5688 \\
CIFAR s1 / epoch 30 & 20 & 1.0096 & 0.6679 & 0.2497 & 0.0502 & 1.0493 \\
\bottomrule
\end{tabularx}
\end{table}

\section{Run provenance and computational cost}
\label{app:provenance}
The initial ImageNet archive contains $5939$ successful outputs, $37$ invalid-anchor aborts and $24$ missing outputs. The latter two categories are not surface-construction failures. The completed set uses $36$ configurations with one substituted anchor and one with two substituted anchors, retaining the target predicted class. They are new boundary configurations in those classes. The $24$ missing runs are completed with their existing anchors. Combining those $61$ outcomes with the $5939$ original successes yields exactly $1000$ accepted configurations covering all $1000$ predicted labels for each model, with zero ceiling-accepted patches in every reported construction. No stronger-repair rescue of the original invalid inputs is claimed.

\begin{table}[ht]
\centering
\small
\caption{Computational cost for the ImageNet study. Times are medians in minutes over $1000$ configurations per model. Component medians need not sum to median total time.}
\label{tab:runtime}
\begin{tabularx}{\linewidth}{@{}>{\raggedright\arraybackslash}X>{\raggedright\arraybackslash}X>{\raggedright\arraybackslash}X>{\raggedright\arraybackslash}X>{\raggedright\arraybackslash}X@{}}
\toprule
Model & Total & Grid & Subdivision & Repair \\
\midrule
ResNet-50 & 24.54 & 13.30 & 7.01 & 6.14 \\
DenseNet-121 & 35.42 & 17.65 & 12.88 & 11.45 \\
EfficientNet-B0 & 21.76 & 11.28 & 7.37 & 6.17 \\
ConvNeXt-Tiny & 13.49 & 9.47 & 1.16 & 0.91 \\
ViT-B/16 & 31.49 & 23.10 & 3.72 & 3.40 \\
Swin-T & 14.33 & 11.10 & 0.31 & 0.27 \\
\bottomrule
\end{tabularx}
\end{table}

Table~\ref{tab:runtime} reports median runtimes. Timing components can overlap, and medians are computed separately, so their sum need not equal median total elapsed time. Comparisons also involve different GPU allocations and batch sizes, so these runtimes should not be interpreted as architecture-independent efficiency rankings. Reproduction requires per-loop identities, final target labels, substituted-anchor provenance and raw acceptance fields; in the archived implementation the ceiling statistic is \texttt{resolution\_leaf\_quads}, which is zero for every accepted run.

\section{Threshold sensitivity and the limits of the archived ablation}
\label{app:thresholds}
The saved threshold archive does not support a blanket claim that all stricter-threshold attempts succeed. Table~\ref{tab:threshold_attempts} retains the repair failures recorded for each threshold. These rows contain unequal configuration sets and are not a matched success-rate experiment.

\begin{table}[ht]
\centering
\small
\caption{Construction outcomes at the non-default grey-RMS thresholds. All recorded attempts, including repair failures, are retained. The configuration sets differ across thresholds, so the acceptance counts do not form a matched comparison.}
\label{tab:threshold_attempts}
\begin{tabularx}{\linewidth}{@{}>{\raggedright\arraybackslash}X>{\raggedright\arraybackslash}X>{\raggedright\arraybackslash}X>{\raggedright\arraybackslash}X@{}}
\toprule
$\tau$ & Attempts & Accepted & Repair failures \\
\midrule
1.0 & 249 & 243 & 6 \\
0.25 & 218 & 214 & 4 \\
0.125 & 121 & 118 & 3 \\
\bottomrule
\end{tabularx}
\end{table}

For a descriptive paired cost comparison, $112$ configurations have successful, ceiling-free records at all four thresholds, including the default $\tau=0.5$ archive. Their anchor paths and target labels agree across thresholds. Table~\ref{tab:threshold_cost} shows that smaller thresholds require more quads in this selected subset. Because selection requires success at every threshold, this table cannot estimate success probability or justify excluding the failures above. These archived runs use the $50$-iteration repair budget.

\begin{table}[ht]
\centering
\small
\caption{Threshold dependence of construction effort for $112$ configurations accepted at all four thresholds. Selection on acceptance makes this a paired cost comparison rather than an estimate of acceptance probability.}
\label{tab:threshold_cost}
\begin{tabularx}{\linewidth}{@{}>{\raggedright\arraybackslash}X>{\raggedright\arraybackslash}X>{\raggedright\arraybackslash}X>{\raggedright\arraybackslash}X>{\raggedright\arraybackslash}X@{}}
\toprule
$\tau$ & Configurations & Mean quads & Mean vertices & Maximum depth \\
\midrule
1.0 & 112 & 820.4 & 924.7 & 8 \\
0.5 & 112 & 2635.5 & 2881.0 & 9 \\
0.25 & 112 & 8899.4 & 9581.1 & 10 \\
0.125 & 112 & 29513.6 & 31799.7 & 11 \\
\bottomrule
\end{tabularx}
\end{table}

\end{document}